\documentclass{article}

\usepackage[preprint, nonatbib]{neurips_2023}

\usepackage[utf8]{inputenc} 
\usepackage[T1]{fontenc}    
\usepackage{hyperref}       
\usepackage{url}            
\usepackage{booktabs}       
\usepackage{amsfonts}       
\usepackage{nicefrac}       
\usepackage{microtype}      
\usepackage{xcolor}         
\usepackage[toc,page,header]{appendix}
\usepackage{wrapfig}
\usepackage{tikz}

\usepackage{subcaption}
\usepackage{marvosym}

\usetikzlibrary{bayesnet, calc, arrows.meta, bending}
\usetikzlibrary{decorations.pathreplacing,angles,quotes,shapes.multipart, calligraphy}

\usepackage{amsmath}

\usepackage[inline]{enumitem}

\usepackage{amssymb}
\usepackage{mathtools}
\usepackage{amsthm}
\usepackage[capitalize,noabbrev]{cleveref}
\usepackage{minitoc}

\usepackage{fontawesome5}
\usepackage{graphicx}

\theoremstyle{plain}
\newtheorem{theorem}{Theorem}
\newtheorem{proposition}{Proposition}
\newtheorem{lemma}{Lemma}

\theoremstyle{definition}
\newtheorem{definition}{Definition}

\theoremstyle{remark}
\newtheorem{remark}{Remark}
\usepackage{printlen}

\usepackage{algorithm}
\usepackage[noend]{algpseudocode}
\usepackage{booktabs}
\usepackage{multirow}
\newcommand{\ffrac}[2]{\ensuremath{\frac{\displaystyle #1}{\displaystyle #2}}}

\newcommand{\gt}{\ensuremath >}

\title{A Causal Ordering Prior for \\ Unsupervised Representation Learning}

\author{%
  Avinash Kori $^{*,\ddagger}$\\
  \texttt{a.kori21@ic.ac.uk} \\
  \And 
  Pedro Sanchez $^{*,\dagger}$\\
  \texttt{pedro.sanchez@ed.ac.uk} \\
  \And
  Konstantinos Vilouras  $^\dagger$\\
  \texttt{konstantinos.vilouras@ed.ac.uk} \\
  \And
  Ben Glocker $^\ddagger$\\
  \texttt{b.glocker@ic.ac.uk} \\
  \And 
  Sotirios A. Tsaftaris $^\dagger$\\
  \texttt{s.tsaftaris@ed.ac.uk} \\
  \AND
  $^*$ {Joint first authors}\\
  $^\dagger$ {School of Engineering, University of Edinburgh} \\
  $^\ddagger$ {Department of Computing, Imperial College London}\\
  }

\usepackage{amsmath,amsfonts,bm}

\def\eqref#1{equation~\ref{#1}}

\def\1{\bm{1}}

\def\rvc{{\mathbf{c}}}

\def\rvf{{\mathbf{f}}}

\def\rvu{{\mathbf{i}}}

\def\rvp{{\mathbf{p}}}

\def\rvu{{\mathbf{u}}}
\def\rvv{{\mathbf{v}}}

\def\rvx{{\mathbf{x}}}

\def\rvz{{\mathbf{z}}}

\def\ervx{{\textnormal{x}}}

\def\ervz{{\textnormal{z}}}

\def\rmX{{\mathbf{X}}}

\def\rmZ{{\mathbf{Z}}}

\DeclareMathAlphabet{\mathsfit}{\encodingdefault}{\sfdefault}{m}{sl}
\SetMathAlphabet{\mathsfit}{bold}{\encodingdefault}{\sfdefault}{bx}{n}

\def\gG{{\mathcal{G}}}

\newcommand{\E}{\mathbb{E}}

\begin{document}

\doparttoc 
\faketableofcontents 

\maketitle

\begin{abstract}
  Unsupervised representation learning with variational inference relies heavily on independence assumptions over latent variables. Causal representation learning (CRL), however, argues that factors of variation in a dataset are, in fact, causally related. Allowing latent variables to be correlated, as a consequence of causal relationships, is more realistic and generalisable. So far, provably identifiable methods rely on: auxiliary information, weak labels, and interventional or even counterfactual data. Inspired by causal discovery with functional causal models, we propose a fully unsupervised representation learning method that considers a data generation process with a latent additive noise model (ANM). We encourage the latent space to follow a causal ordering via loss function based on the Hessian of the latent distribution.
\end{abstract}

\section{Introduction}


The objective of extracting meaningful representations from unlabelled data is a longstanding pursuit in the field of deep learning \cite{bengio2013representation}. Conventionally, methods of unsupervised representation learning have concentrated on unveiling statistically independent latent variables \cite{higgins2017betavae,NIPS2016infoGAN,pmlr-v139-trauble21a,LIU20221disentangled,Higgins2022symmetry}, demonstrating appreciable success in synthetic benchmarks and datasets where generation parameters can be carefully manipulated \cite{locatello2019challenging}. However, it is essential to acknowledge the differences between controlled environments and real-world scenarios. In the latter, the factors contributing to data variation are often intertwined within causal relationships. Therefore, it is not merely advantageous but imperative to integrate causal understanding into the process of learning representations \cite{Bernhard2021towardCRL}, which can improve the models from a generalisation, and interpretability, viewpoint.

The main challenge in learning meaningful and disentangled latent representations is identifiability, 
i.e.\ ensuring the true distribution of a data generation process can be learned (up to a simple transformation, given the inherent limitation that we can never observe the hidden latent factors from observational data alone), implying the model to be injective (one-to-one mapping) onto the observed distribution. Identifiability ensures that if an estimation method perfectly fits the data distribution, the learned parameters will correspond to the true generative model.
For example, discovering independent sources of variation which are observed via a nonlinear mixing function is impossible \cite{hyvarinen1999nonlinear}. This established result from the nonlinear ICA literature has been replicated for disentangled representation learning with variational autoencoders \cite{locatello2019challenging}.

Representation learning becomes identifiable when non-i.i.d.\ (independent and identically distributed) samples from a given data generation process are considered \cite{khemakhem2020variational,hyvärinen2023identifiability}. For instance, temporal contrastive learning \cite{Aapo2016tcl} and iVAE \cite{khemakhem2020variational} can provably ensure identifiability by utilising knowledge of auxiliary information. Indeed, \cite{khemakhem2020variational} develops a comprehensive proof that generative models become identifiable when variables in the latent space are conditionally independent, given the auxiliary information. Conditional independence given external information allows variables to be dependent (or correlated) \cite{Khemakhem2020_ice}, which is more realistic. Further reinforcing the notion of dependence between latent variables, the identifiability of unsupervised representations can be proven by assuming a latent space to follow a Gaussian Mixture Model (GMM) and an injective decoder \cite{kivva2022identifiability}. Any distribution can be approximated by a mixture model with sufficiently many components, including distributions following a causal model. In fact, \cite{kivva2022identifiability} assumes that latent variables are conditionally independent, given a component of the mixture model. The mixture component can correspond to using a ``learned'' auxiliary variable \cite{willetts2021don}, bridging the gap with \cite{khemakhem2020variational}.

These works \cite{Aapo2016tcl,khemakhem2020variational,Khemakhem2020_ice,willetts2021don,hyvärinen2023identifiability} on identifiable representation learning from observational data do not consider latent causal structure. They build up, however, a theory around identifiable representation learning which allows arbitrary distribution encoding statistical dependencies in latent variables. Discovering the dependency structure in the latent space is at the core of causal representation learning (CRL) \cite{Bernhard2021towardCRL} via the \textit{common cause principle}\footnote{``If two observables $X$ and $Y$ are statistically dependent, then there exists a variable $Z$ that causally influences both and explains all the dependence in the sense of making them independent when conditioned on $Z$. As a special case, $Z$ can coincide with $X$ or $Y$.''} \cite{Reichenbach1956ccp}. Learning causally related variables enable 
\begin{enumerate*}[label=(\roman*)]
    \item robustness to distribution shifts via the independent causal mechanism (ICM) principle;
    \item better generalisation, e.g.\ in transfer learning settings;
    \item answering causal queries, i.e.\ estimation of interventional and counterfactual distributions.
\end{enumerate*}
Previous work on CRL, however, utilise data from interventional \cite{ahuja2022interventional,varici2023score} or counterfactual (pre- and post-intervention) \cite{locatello2020weakly,brehmer2022weakly,lippe2022citris} distributions for learning identifiable causal representations. 

In this work, we bridge the gap between identifiable representation learning from observational data and CRL by using functional constraints (which are very common in the causal discovery \cite{Peters2017elements} literature). We propose the first (to the best of our knowledge) method for unsupervised CRL under some data and model assumptions. This can be done by assuming a data generation process in which the latent space adheres to an additive noise model (ANM) and applies an injective nonlinear mapping to generate observational data. The main \textbf{contributions} in this work include
\begin{enumerate*}[label=(\roman*)]
    \item Based on the universal approximation capabilities of GMMs, we show that models with a latent ANM prior are identifiable to block diagonal transformation; and
    \item We propose an estimation method that encourages the latent space to follow an ANM by leveraging asymmetries in the learned latent distribution.
\end{enumerate*}
More specifically, the latent distribution's second-order derivatives (Hessian) can be incorporated into a loss function that promotes latent ordering. We term models trained with the proposed estimation method as \textsc{coVAE} (causally ordered Variational AutoEncoders).

\begin{figure}[t]
  \centering
  \includegraphics[width=\textwidth]{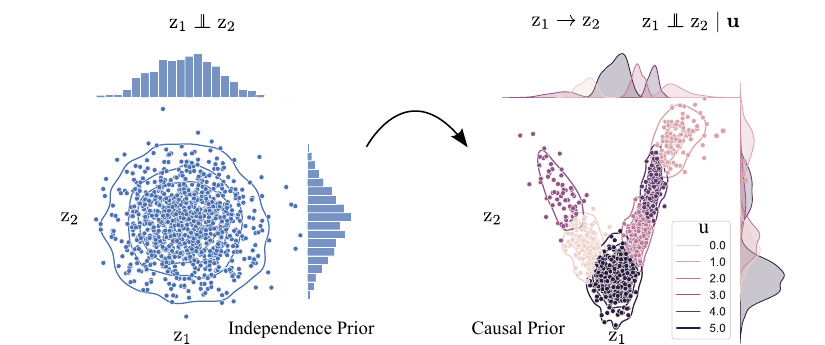}
  \caption{[Left] Independence assumption used in previous work for disentangled representations such as $beta$-VAE and extensions. [Right] We propose to model causally related latent variables. CRL is made possible by using a mixture model in the latent space which approximates an structural causal model (SCM) with functional constraints. $\ervz_1, \ervz_2$ are latent variables, and $\rvu$ correspond to mixture components.}
  \label{fig:identifiability_figure_overview}
\end{figure}

\section{Related Works}
\paragraph{Disentangled Representation Learning.~}
Early efforts on unsupervised representation learning focused on the Variational Autoencoder framework \cite{kingma2013auto}. $\beta$-VAE \cite{higgins2017betavae} and extensions \cite{pmlr-v80-kim18b,eastwood2018a,pmlr-v97-mathieu19a} rely on independence assumptions between latent variables to learn disentangled representations \cite{LIU20221disentangled,Higgins2022symmetry}. Despite showing some success, there is a lack of theory around the identifiability of independent representations. In fact, learning independent (disentangled) representations from i.i.d.\ data in an unsupervised manner is provably impossible \cite{hyvarinen1999nonlinear,locatello2019challenging}.

\paragraph{Representation Learning with Auxiliary Information.~}
A line of work based on nonlinear ICA leverages auxiliary information to learn identifiable models. \cite{khemakhem2020variational} derive a more general proof of identifiability using the concept of conditional independence given auxiliary variables. An extension of nonlinear ICA, called Independently Modulated Component Analysis (IMCA) was proposed in \cite{Khemakhem2020_ice}, where the components are allowed to be dependent. On the contrary, \cite{kivva2022identifiability} prove that identifiability of deep generative models can also be achieved without auxiliary information by considering a GMM prior in the latent space. In the same line, empirical results in \cite{willetts2021don} show that the GMM prior assumption is as efficient as utilising auxiliary information in terms of learning stability (latents learned for different training seeds are correlated). 

\paragraph{Causal Representation Learning.~} 
Following the \textit{common cause principle} \cite{Reichenbach1956ccp}, causal relationships between variables also imply statistical dependencies. Recent works have shown that it is possible to model causal relationships given access to either interventional or non-i.i.d.\ data. To this end, the method in \cite{ahuja2022interventional} uses an injective polynomial decoder and the overall model is trained on both observational and interventional data. Similarly, \cite{varici2023score} consider the case of an injective linear decoder and directly optimize the score function of the distribution (in both the latent and observation space). In \cite{locatello2020weakly} a setting where observations are collected before and after unknown interventions (i.e.\ counterfactual data) is introduced, while \cite{brehmer2022weakly} extends this idea to causal graphs of higher complexity. Under the non-iid scenario, \cite{lippe2022citris} focuses on extracting causal factors from spatiotemporal data by performing interventions across different time steps. There also exist works that assume some level of supervision, i.e.\ having access to ground-truth causal factors. \cite{shen2022weakly} propose a method based on the GAN framework where the prior follows a nonlinear Structural Causal Model (SCM). Others \cite{yang2021causalvae} instead model exogenous noise directly, which is then mapped to causal latent variables via a linear SCM. Table \ref{tab:pretraining_cond} describes data and latent space assumptions of previously existing models in comparison to the proposed method.

\begin{table}[h]
    \centering
    \caption{Comparison of assumptions for identifiability proofs. We classify methods by type of training data: observational (\textit{obs}), interventional (\textit{int}) or counterfactual (\textit{count}); and latent assumptions: independent (\textit{ind}), conditionally independent (\textit{cond ind}), with auxiliary information (\textit{aux}) or structural causal model (\textit{SCM}).}
    \begin{tabular}{ lll } 
     \toprule
     Method & Data & Latents   \\ 
     \midrule
     \textsc{Ada-GVAE} \cite{locatello2020weakly} & count & ind \\
     \textsc{iVAE} \cite{khemakhem2020variational} & obs + aux & cond ind $|$ aux\\
     \textsc{VaDE} \cite{ijcai2017Vade,willetts2021don},  \textsc{MFC-VAE} \cite{falck2021multi,kivva2022identifiability} & obs & cond ind $|$ learned aux \\
     \textsc{CausalVAE} \cite{yang2021causalvae}, \textsc{DEAR} \cite{shen2022weakly} & obs + aux & SCM \\
     \cite{ahuja2022interventional}, \cite{varici2023score} & int & SCM \\
     \textsc{ILCM} \cite{brehmer2022weakly}, \textsc{CITRIS} \cite{lippe2022citris} & count & SCM \\
     Ours (\textsc{coVAE}) & obs & SCM (ANM)\\
     \bottomrule
    \end{tabular}
    \label{tab:pretraining_cond}
\end{table}

\section{Identifiability of Latent Additive Noise Models}

A key challenge in unsupervised representation learning is identifiability. The intuition is that if two parameters result in an identical distribution of observations, then they must be equivalent in order to ensure model identifiability. Note that identifiability is the property of the data generation process, and not of the estimation method. Model identifiability is important because it gives theoretical guarantees that an estimation method is capable of learning the true variables that generated the observed data. Therefore, we first define our model assumptions, show identifiability results and leave the description of the estimation method for the next section. In this section, we define \emph{and} distinguish between the different forms of identifiability and theoretically show that stronger forms of identifiability can be guaranteed when the latent variables are causally ordered.   

\subsection{Preliminaries}
We assume the data generation process maps a latent space $\rvz$, following a structural causal model (SCM), to an observational space $\rvx$ as
\begin{align}
     \rvx = \rvf_o(\rvz) + \mathbf{\epsilon}_x , && \mathbb{P}(\rvz) = \prod_{i} \mathbb{P}(\ervz_i \mid \mathbf{pa}(\ervz_i)).
    \label{eqn:data_generation_process}
\end{align}
$\rvf_o: \mathbb{R}^d \rightarrow \mathbb{R}^o$ is a non-linear injective mapping (or mixing function), $d$ is the number of latent variables and $o = |\mathcal{O}| \geq d$. $\mathbb{P}(\rvz)$ is a distribution entailed by a SCM following a directed acyclic graph (DAG) $\gG$, containing $d$ nodes, which describes the true causal structure of the latent. $\mathbf{pa}(\ervz_i)$ are the parents of $\ervz_i$ in $\gG$.

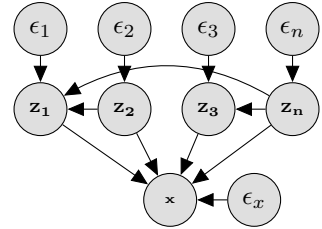
\begin{wrapfigure}{r}{0.35\textwidth}
    \centering
    \begin{tikzpicture}
        \node[obs, minimum size=20pt] (eps1) {$\epsilon_1$};
        \node[obs, below=0.35cm of eps1, font=\fontsize{5}{5}\selectfont] (z1) {$\mathbf{\rmZ_1}$};
        
        \node[obs, right=0.4cm of eps1] (eps2) {$\mathbf{\epsilon}_2$};
        \node[obs, below=0.35cm of eps2, font=\fontsize{5}{5}\selectfont] (z2) {$\mathbf{\rmZ_2}$};

        \node[obs, right=0.4cm of eps2] (eps3) {$\mathbf{\epsilon}_3$};
        \node[obs, below=0.35cm of eps3, font=\fontsize{5}{5}\selectfont] (z3) {$\mathbf{\rmZ_3}$};

        \node[obs, right=0.4cm of eps3] (epsn) {$\mathbf{\epsilon}_n$};
        \node[obs, below=0.35cm of epsn, font=\fontsize{5}{5}\selectfont] (zn) {$\mathbf{\rmZ_n}$};

        \node[obs, below right=0.7cm and 0.1cm of z2, font=\fontsize{3}{3}\selectfont] (x) {$\mathbf{\rvx}$};

        \node[obs, right=0.4cm of x] (epsx) {$\mathbf{\epsilon}_x$};
        
        \path (eps1) edge [->] (z1)
        (eps2) edge [->] (z2)
        (eps3) edge [->] (z3)
        (epsn) edge [->] (zn)
        (z2) edge [->] (z1)
        (zn) edge [->] (z3)
        (zn) edge [bend right,->] (z1)
        (z1) edge [->] (x)
        (z2) edge [->] (x)
        (z3) edge [->] (x)
        (zn) edge [->] (x)
        (epsx) edge [->] (x);

    \end{tikzpicture}
    \caption{Data generation process with a latent SCM (endogenous and exogenous variables) causing an observation space.}
\end{wrapfigure}

\textbf{Additive Noise Models.~} We assume that the latent SCM consists of a collection of assignments following an additive noise model (ANM) $\ervz_i \coloneqq f_i(\mathbf{pa}(\ervz_i)) + \epsilon_i$. $\epsilon_i$ is a noise term independent of $\ervx_i$, also called exogenous noise. $\epsilon_i$ are i.i.d.~from a smooth distribution $\mathbb{P}^\epsilon$. When using an ANM assumption over $\rvz$, the latent distribution in \ref{eqn:data_generation_process} becomes
\begin{equation}
    \mathbb{P}(\ervz) = \prod_{i} \mathbb{P}(\ervz_i \mid \mathbf{pa}(\ervz_i)) =  \prod_{i} \mathbb{P}^\epsilon(\ervz_i - f_i(\mathbf{pa}(\ervz_i))).
\end{equation}
This assumption is particularly important to demonstrate guarantees on stronger forms of identifiability.
Assuming a functional form for the causal mechanism between variables, such as ANMs \cite{Hoyer2008ANM,peters14acANM}, is an established method for identifying causal relationships \cite{Peters2017elements,Glymour2019discovery} due to asymmetries in the joint distribution. Moreover, the ANM assumption has been shown to perform well on real benchmarks from various domains such as meteorology, biology, medicine, engineering and economy \cite{mooij2016distinguishing}, for the task of causal discovery. 

\textbf{Causal Ordering.~} Since we assume $\gG$ to be a DAG, there is a non-unique permutation $\tau$ of $d$ nodes such that a given node always appears first in the list compared to its descendants. Formally,  $\tau_i < \tau_j \iff j \in  \mathbf{de}(\ervz_i)$ where $\mathbf{de}(\ervz_i)$ are the descendants of $\ervz_i$ in $\gG$ (Appendix B in \cite{Peters2017elements}).



\subsection{Identifiability Equivalence}

The exact definition of model identifiability can be too restrictive. In reality, identifying a representation up to a simple transformation is enough. Therefore, we now formally define identifiability \ref{dfn:strongidentifiability} and its weaker forms, which guarantee identifiability up to affine transformation \ref{dfn:Aequivalence}, permutation and scaling \ref{dfn:Pequivalence}, and block diagonal and scaling transformations \ref{dfn:Iequivalence}.
In the case of an ANM data generating process, \cite{peters2014causal} demonstrates the identifiability of models with only observational data; further, \cite{rolland2022score} discuss the identifiability of these models under data \emph{score} functions. However, they do not discuss the identifiability of latent ANM models.

In this section, we define and make a distinction between different forms of identifiabilities and theoretically show that stronger forms of identifiability can be guaranteed when latent variables are causally ordered.  

\begin{definition}(Strong Identifiability)
For parameter domain $\Theta$ and equivalence relation $\sim$ on $\Theta$, the considered model is $\sim$-identifiable if equation \ref{eqn:strongidentifiability} is satisfied.
\begin{equation}
    \mathbb{P}_{\theta_1}(\rvx) = \mathbb{P}_{\theta_2}(\rvx), \Rightarrow \theta_1 \sim \theta_2.
    \label{eqn:strongidentifiability}
\end{equation}
\label{dfn:strongidentifiability}
\end{definition}
\begin{remark}
According to \cite{khemakhem2020variational}, strong model identifiability makes the latent space $\mathbb{P}(\rvz)$ identifiable. 
\end{remark}

\begin{definition}(Affine Equivalence, $\sim_A$)
For $\theta = \{\mathbf{f}, \mathbf{p}\}$ a set of parameters corresponding to the mixing function and prior, the affine equivalence relation $\sim_A$ on $\Theta$ is defined as:
\begin{equation}
    (\mathbf{f}, \mathbf{p}) \sim_A (\mathbf{\tilde{f}}, \mathbf{\tilde{p}}) \iff 
\exists \quad  \mathbf{A} \in \mathbb{R}^{n\times n}, \mathbf{c} \in \mathbb{R}^n \quad s.t. \quad  \mathbf{f}^{-1}(\rvx) = \mathbf{A} \mathbf{\tilde{f}^{-1}(\rvx) + \mathbf{c}}, \forall \rvx \in \mathcal{O}.
\label{eqn:Aequivalence}
\end{equation}
where $\mathbf{A}$ is an invertible matrix and $\mathcal{O}$ is an observational data space. 
\label{dfn:Aequivalence}
\end{definition}

\begin{remark}
$\sim_A$ states that the images of $\mathbf{f}^{-1}$ and $\mathbf{\tilde{f}^{-1}}$ are related by an affine transformation.
\end{remark}

\begin{definition}(Permutation Equivalence, $\sim_P$)
For $\theta = \{\mathbf{f}, \mathbf{p}\}$ a set of parameters corresponding to the mixing function and prior, the permutation equivalence relation $\sim_P$ on $\Theta$ is defined as by:
\begin{equation}
    (\mathbf{f}, \mathbf{p}) \sim_P (\mathbf{\tilde{f}}, \mathbf{\tilde{p}}) \iff 
\exists \quad  \mathbf{P} \in \mathbb{R}^{n\times n}, \mathbf{c} \in \mathbb{R}^n \quad s.t. \quad  \mathbf{f}^{-1}(\rvx) = \mathbf{P} \mathbf{\tilde{f}^{-1}(\rvx) + \mathbf{c}}, \forall \rvx \in \mathcal{O}.
\label{eqn:Pequivalence}
\end{equation}
where $\mathbf{P}$ is a block permutation matrix and $\mathcal{O}$ is an observational data space. 
\label{dfn:Pequivalence}
\end{definition}

\begin{remark}
$\sim_P$ states that the images of $\mathbf{f}^{-1}$ and $\mathbf{\tilde{f}^{-1}}$ are related by rotation, scaling, and translation.
\end{remark}

\begin{definition}(Block Diagonal Equivalence, $\sim_D$)
For $\theta = \{\mathbf{f}, \mathbf{p}\}$ a set of parameters corresponding to the mixing function and prior, the identity equivalence relation $\sim_D$ on $\Theta$ is defined as by:
\begin{equation}
    (\mathbf{f}, \mathbf{p}) \sim_D (\mathbf{\tilde{f}}, \mathbf{\tilde{p}}) \iff 
\exists \quad  \mathbf{D}, \mathbf{c} \quad s.t. \quad  \mathbf{f}^{-1}(\rvx) = \mathbf{D} \mathbf{\tilde{f}^{-1}(\rvx) + \mathbf{c}}, \forall \rvx \in \mathcal{O}.
\label{eqn:Dequivalence}
\end{equation}

where $\mathbf{D}$ is a block diagonal matrix, $\mathbf{c} \in \mathbb{R}^d$ is a shift vector, and $\mathcal{O}$ is an observational data space.
\label{dfn:Iequivalence}
\end{definition}

\begin{remark}
$\sim_D$ states that the images of $\mathbf{f}^{-1}$ and $\mathbf{\tilde{f}^{-1}}$ are related just by translation and scaling.
\end{remark}

\subsection{Identifiability of Latent ANMs} 

\paragraph{Universal Approximation of GMMs.~}
Assuming the data generating process is an affine or piece-wise affine function, GMMs with a sufficient amount of components can model any densities in the limiting case \cite{nguyen2019approximations}, which in turn breaks the symmetry in the latent space behaving like auxiliary information in iVAE \cite{willetts2021don, kivva2022identifiability}.
In light of this, we model our latent distribution $\mathbb{P}(\ervz) = \prod_{i} \mathbb{P}^\epsilon(\ervz_i - f_i(\mathbf{pa}(\ervz_i))) = \sum_{j=1}^J \pi_j \mathcal{N}(\mu_j, \Sigma_j)$ as a mixture of densities. 

\begin{theorem}(Identifiability of $\rvz$ under $\gG$)
Let $\rvf_o, \tilde{\rvf}_o$, satisfying injectivity assumption with $y \sim \mathbb{P}(\rvz), y' \sim  \tilde{\mathbb{P}}(\rvz)$, where $\mathbb{P}, \tilde{\mathbb{P}}$ follow the same causal graph $\gG$. Suppose $\rvf_o(y)$ and $\tilde{\rvf}_o(y')$ are equally distributed, then, $\mathbb{P}(\rvz) \sim \tilde{\mathbb{P}}(\rvz)$.
\label{thm:latentidentifiability}
\end{theorem}


\begin{remark}
This theorem is similar to, but goes beyond, Theorem E.1 in \cite{kivva2022identifiability}. We show equivalence up to $\sim$ rather than $\sim_P$, given that the latent variables are constrained with respect to some causal graph (with all conditional independencies).
\end{remark}

The proof is detailed in the appendix. The main outline of this proof includes showing that, under the constrain that the latent distribution respects the same causal graph $\gG$, the block permutation matrix (in Theorem E.1 of \cite{kivva2022identifiability}) can be reduced to a diagonal matrix. Similar to \cite{kivva2022identifiability} we approximate the posterior distribution using GMMs.

\begin{lemma}(Identifiability of $\rvz$ under causal ordering) In the case when only causal ordering is known, the strong identifiability in theorem \ref{thm:latentidentifiability} reduces to block diagonal identifiability ($\sim_D$).
\label{lemma:orderedlatentidentifiability}
\end{lemma}

\begin{remark}
Given the fact that constraining latent variables based on the complete causal graph may not be feasible, the lemma relaxes this constraint to enforce causal ordering, which guarantees $\sim_D$ identifiability. In section \ref{sec:estimation}, we show how to achieve causal ordering in the latent space.
\end{remark}

\begin{theorem}(Model Identifiability)
Let $\rvf_o, \tilde{\rvf}_o$, satisfy the injectivity assumption with $y \sim \mathbb{P}(\rvz), y' \sim  \tilde{\mathbb{P}}(\rvz)$, where $\mathbb{P}, \tilde{\mathbb{P}}$ follow the same causal graph $\gG$ and let $\mathcal{D} \subseteq \mathbb{R}^o$, where $o = |\mathcal{O}|$ such that $\rvf_o, \tilde{\rvf}_o$ are injective on to $\mathcal{D}$. 
Suppose $\rvf_o(y)$ and $\tilde{\rvf}_o(y')$ are equally distributed, then, $\rvf_o(\rvz) = \tilde{\rvf}_o(\rvz)$.

\label{thm:modelidentifiability}
\end{theorem}

\begin{remark}
This theorem is similar to, but goes beyond Theorem D.4 in \cite{kivva2022identifiability}. We show equivalence up to $\sim$ rather than $\sim_A$, given that the latent variables are constrained with respect to some causal graph (with all conditional independencies).
\end{remark}

We detail the proof in the appendix. Similar to the proof of Theorem \ref{thm:latentidentifiability}, we use GMMs to model our posterior distribution. The main component of the proof is to reduce affine transformation in Theorem D.4 \cite{kivva2022identifiability}  to an identity transformation.

\begin{lemma}(Model identifiability under causal ordering) In the case when latent variables follow a particular causal ordering $\tau$ rather than the entire causal graph $\gG$, there exists a block diagonal transformation $\mathbf{D}$ such that $\rvf_o(\rvz) = (\tilde{\rvf}_o \circ \mathbf{D})(\rvz)$.

\label{lemma:orderedmodelidentifiability}
\end{lemma}
\section{Estimation}
\label{sec:estimation}

We now derive an estimation procedure for learning the data generation process in equation \ref{eqn:data_generation_process}. The findings of the previous section show that a data generation process with an ANM in the latent space is identifiable if the causal graph (or causal ordering) is known. Therefore, we proceed to define a loss function that will ensure that the latent space is causally ordered. Then, we describe a variational inference estimation method which models latent variables using a GMM.

\subsection{Causal Ordering Loss}

In causal representation learning, the goal is to learn causal variables from observations without information about the causal structure. However, there is always a causal ordering associated with a DAG. It is well known in the causal discovery literature that a complete causal graph is not identifiable from observational data without extra assumptions. If the functional form of the causal mechanism is assumed to be an ANM, causal directions become identifiable due to asymmetries. Interestingly, previous works on causal discovery \cite{rolland2022score,sanchez2023diffusion} explore a property of the distribution of ANMs to find a causal ordering. Here, we use the same property to enforce causal ordering instead of discovering it. 

Enforcing causal ordering allows us to approximate the assumption of known causal ordering from Lemma \ref{lemma:orderedlatentidentifiability}. We use this property as a loss function for learning the latent representations.
The property is based on the Jacobian of an ANM distribution's score function. Firstly, let the latent distribution be $\mathbb{P}(\rvz)$ which follows an ANM and $\mathbb{P}^{\epsilon}$ be any quadratic exponential noise prior (e.g.\ Gaussian-like) \cite{rolland2022score,sanchez2023diffusion}. We can express its score function as
\begin{equation}
    \nabla_{\ervz_{i}}\log \mathbb{P}(\rvz) = \frac{\partial \log \mathbb{P}^\epsilon(\ervz_i - f_i(\mathbf{pa}(\ervz_i)))}{\partial \ervz_i} - \sum_{j \in \mathbf{ch}(\ervz_i)} \frac{\partial f_j}{\partial \ervz_i} \frac{\partial \log \mathbb{P}^\epsilon(\ervz_j - f_j(\mathbf{pa}(\ervz_j)))}{\partial \ervz_i}.
\label{eqn:score_anm}
\end{equation}
Based on the above formalism 
it can be derived that $\nabla_{\rvz_{i}}^2 \log \mathbb{P}(\rvz) = a \iff \rvz_{i}$ is a leaf node, where $a$ is some constant and $\nabla_{\rvz_{i}}^2 \log \mathbb{P}(\rvz)$ is $i^{th}$ diagonal element of the distribution's Hessian. 
\begin{proposition}
Assuming that $\mathbb{P}(\rvz)$ follows an ANM and let $H_{var}^{i}(\rvz) = \mathrm{var} \Big(\nabla^2_{\ervz_{i}} \log \mathbb{P}(\rvz)\Big)$. The latent space $\rvz$ can be causally ordered by minimising the causal ordering loss defined as
\begin{equation}
    \mathcal{L}_{order} =  -\sum_i^{d-1} \log \frac{{H_{var}^{i}(\ervz_i, \dots, \ervz_d)}^{-1}}{\sum_{j = i}^d {H_{var}^{j}(\ervz_i, \dots, \ervz_d)}^{-1}}
    \label{eqn:causal_loss}
\end{equation}   
\label{prop:latentorganisation}
\end{proposition}


\begin{proof}
The proof directly extends from analysing equation \ref{eqn:score_anm}. As described in \cite{rolland2022score}, the minimum variance in the latent log-likelihood's hessian corresponds to a leaf node.
The loss term $\mathcal{L}_{order}$ is minimum if, and only if, the nodes at position $i$ are leaves. 
We show this by contradiction; without loss of generality, consider the random latent order $\tau$, s.t.\ $\tau_i \neq i$, then
$H_{var}^{0}(\rvz) \geq \epsilon \Rightarrow \mathcal{L}_{order} \gt 0$.
Based on the above expression $\mathcal{L}_{order} \rightarrow 0, \iff  \tau_i = i$, where $\tau_i$ correspond to true causal order. 
It is important to note that as the representations are learned end-to-end, enforcing this loss would organise the latent order to follow the sorted true causal ordering. 
\end{proof}

\textbf{Hessian Estimation.~} To compute $H_{var}^{i}(\rvz)$, we approximate the score's Jacobian (Hessian) with Stein kernel estimators \cite{li2017gradient} as described in \cite{rolland2022score}: 
\begin{equation}
    \mathbf{J}^{Stein}  = -\mathrm{diag}(\mathbf{G}^{Stein}(\mathbf{G}^{Stein})^T) + (\mathbf{K} + \eta \mathbf{I})^{-1}\langle \nabla^2_{diag}, \mathbf{K}\rangle 
    \label{eqn:stein_estimators}
\end{equation}
Where $\mathbf{G}^{Stein} = -(\mathbf{K} + \eta \mathbf{I})^{-1}\langle \nabla, \mathbf{K}\rangle$ is the Stein gradient estimator \cite{li2017gradient}, $\mathbf{K}$ is the median kernel, $\mathbf{I}$ is the Identity matrix, and $\langle a, b \rangle$ correspond to applying operation $a$ on $b$ element-wise. 
The final algorithm for computing $\mathcal{L}_{order}$ is described in Alg.\ \ref{alg:top_loss}.

\begin{algorithm}[ht]
\caption{Compute topological loss ($\mathcal{L}_{order}$)} 
\label{algo:topologicalloss}
    \begin{algorithmic}[1]
        \State \textbf{Initialize:} $ \mathcal{L}_{order} = 0$
        \State \textbf{Given:} $ \rvz = \rvf^{-1}(\rvx)$
        
        \State \textbf{for} $i=0, \dots, d-1$
        \State \quad $\tilde{\rvz} = \rvz[i:]$ 
        \State \quad $\rvv = \mathrm{var}\left( \mathbf{J}^{Stein}(\tilde{\rvz}) \right)$ \Comment{Compute variance of a Jacobian of a score}   

        \State \quad $\tilde{\rvv} = \mathrm{softmax}(- \log \rvv)$ \Comment{Smallest variance $\rightarrow$ highest $\tilde{\rvv}$ }

        \State \quad $\mathcal{L}_{order} += \text{BCE}(\tilde{\rvv},[1,0 \dots 0])$ \Comment{First element should have smallest variance}

        \State \textbf{return} $\mathcal{L}_{order}$
    \end{algorithmic}
    \label{alg:top_loss}
\end{algorithm}

\subsection{Variational Inference}

We are now interested in modelling a latent space with an arbitrarily complex distribution based on an ANM using the deep variational framework. That is, learning a posterior distribution that can approximate the ANM prior $\mathbb{P}(\rvz)$ given a sample from the observational distribution. A multivariate diagonal Gaussian prior cannot model these distributions. Therefore, we consider a prior following a GMM, following established literature \cite{jiang2016variational,Johnson2016composing,falck2021multi}, which is proven to be identifiable and have universal approximation capabilities \cite{kivva2022identifiability}.

In particular, we utilise the framework from MFC-VAE \cite{falck2021multi}.
We consider the generative model to be $\mathbb{P}(\rvx,\rvz,\rvu) = \mathbb{P}(\rvx\mid\rvz)\mathbb{P}(\rvz\mid\rvu)\mathbb{P}(\rvu)$. MFC-VAE choose a posterior $\mathbb{Q}(\rvu,\rvz\mid\rvx) = \mathbb{Q}(\rvu\mid\rvx)\mathbb{Q}(\rvz\mid\rvx)$, where $\mathbb{Q}(\rvz\mid\rvx)$ is a multivariate Gaussian with diagonal covariance and $\mathbb{Q}(\rvu\mid\rvx)$ a categorical distribution over GMM components. 
Similar to MFC-VAE \cite{falck2021multi},  we consider our inference model as described above, where the mixture components are inferred via prior (as $\mathbb{Q}_{\rvf}(\rvu \mid \rvx) \propto \exp (\mathbb{E}_{\mathbb{Q}_{\rvf}(\rvz \mid \rvx)} \log \mathbb{P}_{\rvp}(\rvu \mid \rvz) )$). 
In this case, the posterior $\mathbb{Q}(\rvu,\rvz\mid\rvx)$ is a GMM and can approximate the prior $\mathbb{P}(\rvz)$ following a ANM. The ELBO for this model is described in Eqn. \ref{eqn:elbo}, where $\mathbb{E}$ is over $\mathbb{Q}(\rvz\mid\rvx)$ distribution.
\begin{equation}
 \mathcal{L}_{ELBO} = - \E
 \log \mathbb{P}(\rvx\mid\rvz) +\mathrm{KL}\Big(\mathbb{Q}(\rvz \mid \rvx) \mid\mid \mathbb{P}(\rvz)\Big) + \E \;
 \mathrm{KL}\Big(\mathbb{Q}(\rvc\mid\rvx) \mid\mid \mathbb{P}(\rvc\mid\rvz)\Big)    
\label{eqn:elbo}
\end{equation}
 




\begin{lemma}(Training Objective)
Based on the proposition \ref{prop:latentorganisation} and lemmas \ref{thm:latentidentifiability} and \ref{thm:modelidentifiability}, models trained with the following objective: $\mathcal{L}_{total} = \mathcal{L}_{ELBO} + \alpha \mathcal{L}_{order}$, where 
 will converge at true latents with $\sim_D$ equivalence.   
\label{lemma:training_objective}
\end{lemma}

\subsection{Neural Network Constraints}
\label{sec:decoderconstraints}

\textbf{Injective Decoder.~} It is common to assume an injective decoder for proving the identifiability of a data generation process \cite{kivva2022identifiability}. When implementing a deep generative model in practice, some constraints in the decoder are necessary to ensure that neural networks are modelling injective functions. We follow similar modelling assumptions of ICE-BeeM \cite{Khemakhem2020_ice}:
\begin{enumerate*}[label=(\roman*)]
    \label{assumption:decodermodelling}
    \item Monotonicity: The latent dimension of the decoder is monotonically increasing, \emph{i.e.,} $d_{l+1} \geq d_l \quad \forall l \in \{0, \dots, L-1 \}$, where $d_l$ corresponds to the feature dimension at layer $l$ and $L$ is the total number of layers in the decoder.
    \item Activation: The activation function after every layer corresponds to LeakyReLU ($\max(0, x) + \alpha \min(0, x), \alpha \in (0, 1)$).
    \item Full rank: All weight matrices $\mathbf{f}_l$ are full row ranked, as the number of rows is greater than or equal to the number of columns. 
    \item Invertible sub-matrix: All weight sub-matrices $\mathbf{f}'_l$ of size $d_l \times d_l$ are invertible.
\end{enumerate*}

\textbf{Discussion:} Proposition \ref{prop:latentorganisation} shows that, given sufficient data and compute,  under the non-linear ANM assumption, latent representations are organised with respect to evidential ordering. Additionally, given the organised latent representations, the causal relationships among the representations can be estimated using conditional independencies similar to \cite{rolland2022score, sanchez2023diffusion, kalisch2007estimating}. We later discuss how latent causal discovery can be achieved.
As previously discussed in equation \ref{eqn:data_generation_process}, it is important to note that we consider all features in $\rvz$ to be direct parents of $\rvx$, thus any indirect cause $y \rightarrow (\rvz_i \in \rvz) \rightarrow \rvx$ cannot be recovered by our approach.

\section{Experiments}

Here, we demonstrate the effectiveness of latent ANM models with topological constraints on both tabular (including a synthetic data generating process) and image (MorphoMNIST and Causal3DIdent) datasets. We compare the proposed model against two baseline methods $\beta$-VAE and MFC-VAE with a single facet on mean correlation coefficient (MCC) and causal ordering divergence (COD).

\subsection{Metrics}

We compute different variants of MCC: (i) across multiple random seeds (MCC-R): measures the stability of the training process given the model; (ii) with respect to ground truth variables (MCC-GT): measures the faithfulness of the estimated latent variables to true latent variables \cite{Khemakhem2020_ice}; and (iii) subset MCC (MCC-SG): in the case when all parents of $\rmX$ are not observed, we measure the faithfulness by considering a subset of latent variables. All three variants are formally described in definition \ref{dfn:mcc}.
As these MCC measures are permutation invariant by nature, to capture the perceived order among latent variables, we also calculate COD, which measures the divergence of the topological order in an estimated causal graph from the causal order, formally defined in equation \ref{eqn:COD}.
In addition, to quantify the injectiveness of the model we compute MIC and RRO defined in \ref{dfn:mic}. 

\begin{definition}(Mean Correlation Coefficient)
We compute the mean correlation coefficient with respect to ground truth (MCC-G) as described in \cite{Khemakhem2020_ice}. MCC-SG and MCC-R are based on MCC-G and are described as:


\begin{equation}
    \textsc{MCC-SG}(\hat{\rvz}, \rvz) =  \max \Big\{ \textsc{MCC-G}(\hat{\rvz}[S_j], \rvz), \quad  \forall j=\{1,\dots, |\mathcal{S}|\}, \quad S = \binom{|\hat{\rvz}|}{|\rvz|}\Big\}
    \label{eqn:mccg}
\end{equation}

\begin{equation}
    \textsc{MCC-R}(\{\hat{\ervz}_0, \dots, \hat{\ervz}_K\}) = \ffrac{1}{K-1} \sum_k \textsc{MCC-G}(\hat{\ervz}_k, \hat{\ervz}_0),
    \label{eqn:mccr}
\end{equation}

where $\hat{\rvz}_k = \mathbf{f}_k^{-1}(\rmX)$, $S$ is the set of all the partition indices of $\hat{z}$ with the size of $|\rvz|$,  $\rvz$ corresponds to the ground truth latent features and $K$ total number of experimental runs.
\label{dfn:mcc}
\end{definition}

\begin{definition}(Causal Order Divergence, COD) Similar to divergence metric in \cite{rolland2022score,sanchez2023diffusion}, we define COD as: 
\begin{equation}
    \mathrm{COD}({\tau}, A) = \sum_{i=0}^d \sum_{j>i}^d A_{ij}
    \label{eqn:COD}
\end{equation}

where ${\tau}=\{0, \dots, d\}$ is the expected order and $A$ is an estimated adjacency graph predicted using the resulting latent space after training.
\label{dfn:COD}
\end{definition}

\begin{definition}(Mean Injectivity Coefficient, MIC)
Based on the network constraints described in section \ref{sec:decoderconstraints}, we compute the MIC to measure the \emph{injectivity} of the model. MIC is formally described as:
\begin{equation}
    \textsc{MIC}(\mathbf{f}) = \min \Big\{ \ffrac{1}{|\mathcal{C}|} \sum_j \ffrac{Rank(\mathbf{f}_i(\mathcal{C}_j)^T)}{ri} \quad \forall i \in \{0, \dots, |\rvf|\}\Big\}
    \label{eqn:mic}
\end{equation}
where, $ci, ri$ correspond to number of columns and rows of $\rvf_i$, with abuse of notation, we use $\mathcal{C} = \binom{ci}{ri}$ as a set of all partitions of column indices with size $ri$, and $|S|$ is the cardinality of set $S$. 
\label{dfn:mic}
\end{definition}

\begin{remark} 
We measure the average row rank ratio $ \mathrm{RRO} = \Big( \frac{1}{L} \sum_l \ffrac{Rank(f_l)}{d_l} \Big)$ and \textsc{MIC} (ref. definition \ref{eqn:mic}) to measure the injectivity of the decoder.
\end{remark}

\subsection{Data Generation}

\textbf{Simulation Data: } To generate the synthetic dataset we first randomly generate a latent causal DAG with $n$ nodes and $e$ edges using \cite{zhang2021gcastle}. We randomly select all the involved structural causal models $f_i$ with an \emph{injective} mapping from $\mathbf{pa}(\ervz_i)$ to $\ervz_i$. Finally, we select an injective random transformation function $\rvf_o$ mapping from latent space $\rvz$ to observational data $\rmX$. In our experiments we generate 2,000 datapoints from \textsc{Syn-2}, \textsc{Syn-15}, and \textsc{Syn-50} processes, where \textsc{Syn-k} correspond to the above data-generating process with latent variable $\rvz \in \mathbb{R}^{k}$ and observational data $\rmX \in \mathbb{R}^{2k}$.

\textbf{Image Datasets: } We further extend our method on imaging datasets, which include MorphoMNIST \cite{castro2019morpho} variants and Causal3DIdent \cite{von2021self}. In the case of MorphoMNIST, we use MorphoMNIST-IT, MorphoMNIST-TI, MorphoMNIST-TS, and MorphoMNIST-TSWI variants where I, T, S, and W
correspond to latent variables $\rvz$ with the semantics of intensity, thickness, slant, and width respectively. We detail all the data-generating processes in Appendix. All the MorphMNIST variants have 60,000 training images and 10,000 testing images. Similarly, Causal3DIdent includes 252,000 training samples and 25,200 test samples that were generated using a fixed causal graph with 10 nodes (more details about this dataset can be found in \cite{von2021self}, Appendix B).

\subsection{Results}    

\begin{wraptable}{r}{0.5\textwidth}
    \caption{MCC and COD results on synthetic datasets with 2, 15, and 50 nodes in the latent space along with imaging datasts MorphoMNIST-IT and MorphoMNIST-TSWI.}
    \resizebox{0.49\textwidth}{!}{
        \begin{tabular}{@{}cccc@{}}
            \toprule
            \multirow{2}{*}{\begin{tabular}[c]{@{}l@{}} \textsc{Methods}($\downarrow$),\\ \textsc{Metrics}($\rightarrow$)\end{tabular}} & \multicolumn{3}{c}{\textsc{Syn-2}} \\ \cmidrule(l){2-4}
             & COD ($\downarrow$) & MCC-R($\uparrow$)   & MCC-G($\uparrow$) \\ \cmidrule(r){1-4}
             VAE  & 0.13 $\pm$ 0.08 &  0.11  & 0.26$\pm$ 0.03  \\
             MFC-VAE & 0.17 $\pm$ 0.09 &  0.14  & 0.35 $\pm$ 0.06 \\
             \textsc{coVAE} & \textbf{0.00} $\pm$ 0.01 &  \textbf{0.62}  & \textbf{0.52} $\pm$ 0.07\\ \cmidrule(r){1-4}
             & \multicolumn{3}{c}{\textsc{Syn-15}} \\ \cmidrule(r){1-4}
             VAE & 1.68 $\pm$ 0.22  &  0.21 &  0.22 $\pm$ 0.02  \\
             MFC-VAE & 1.43 $\pm$ 0.24 & 0.26 & 0.26 $\pm$ 0.03   \\
             \textsc{coVAE} & \textbf{0.03} $\pm$ 0.01 & \textbf{0.42} & \textbf{0.34} $\pm$ 0.03  \\ \cmidrule(r){1-4}
             & \multicolumn{3}{c}{\textsc{Syn-50}} \\ \cmidrule(r){1-4}
             VAE  & 5.53 $\pm$ 0.81 & 0.23 &  0.28 $\pm$ 0.24\\
             MFC-VAE & 5.17 $\pm$ 0.62  &  0.31 &  0.26 $\pm$ 0.01\\
             \textsc{coVAE} & \textbf{0.78} $\pm$ 0.46  & \textbf{0.39} &  \textbf{0.34} $\pm$ 0.02\\ \cmidrule(l){1-4}
             & \multicolumn{3}{c}{\textsc{MorphoMNIST-IT}} \\\cmidrule(l){2-4}
             & COD ($\downarrow$) & MCC-R($\uparrow$)   & MCC-SG($\uparrow$) \\ \cmidrule(r){1-4}
             VAE  & 1.61 $\pm$ 0.44  & 0.29 & 0.23 $\pm$ 0.11\\
             MFC-VAE & 1.04 $\pm$ 0.46  & 0.36 & 0.34 $\pm$ 0.09 \\
             \textsc{coVAE} & \textbf{0.0} & \textbf{0.59} & \textbf{0.47} $\pm$ 0.08 \\ \cmidrule(r){1-4}
             & \multicolumn{3}{c}{\textsc{MorphoMNIST-TSWI}} \\ \cmidrule(r){1-4}
             VAE & 0.81 $\pm$ 0.26 &  0.47 & 0.21 $\pm$ 0.00\\
             MFC-VAE & 1.35 $\pm$ 0.24 & 0.52   & 0.28 $\pm$ 0.04\\
             \textsc{coVAE} & \textbf{0.0} &  \textbf{0.61}  & \textbf{0.31} $\pm$ 0.04\\
             \bottomrule
        \end{tabular}
    }
    \label{tab:results}
\end{wraptable}

In each of our experiments, we adopt a model adhering to the properties delineated in Section \ref{sec:decoderconstraints}. Observations pertaining to MIC and RRO measures suggest that the injectivity of the decoder is predominantly influenced by choice of architecture and the dataset under consideration.

For instance, the MIC for the \textsc{Syn-2}, \textsc{Syn-15}, and \textsc{Syn-50} datasets are recorded as 1.0, 0.68, and 1.0, respectively, while the corresponding RRO values are 0.88, 0.93, and 0.95. To gauge the effectiveness in terms of stability and faithfulness, we tabulated the results concerning MCC-R and MCC-GT metrics for synthetic and image datasets in Table \ref{tab:results}. Here, we employed five random seeds to compute the MCC-R and report the mean and standard deviation across these five runs for COD and MCC-G.
These results, illustrate that given additive noise models in latent space, the proposed loss enforces evidential structure as COD goes to 0 and achieves stronger identifiability which can be inferred from MCC-R and MCC-G values.

Similarly, in the case of imaging datasets for both MorphoMNIST-IT and MorphoMNIST-TSWI we observed MIC of 1.0 and RRO of 0.85, and the resulting MCC-SG (as previously described, in the case of image datasets, all the parents are not observed) and COD measures are described in Table \ref{tab:results}. 
In all our experiments, we observed that topological ordering with respect to the evidential graph is better enforced in \textsc{coVAE} and even in terms of stability and faithfulness of the latent representations, \textsc{coVAE} outperforms VAE and MFC-VAE.  
Additional experiments on other variants of the MorphoMNIST dataset and Causal3DIdent are detailed in the Appendix.

\section{Conclusion}
In this work, we propose the first fully unsupervised causal representation learning method for data 
adhering to ANM by imposing a topological ordering on the latent space that corresponds to the underlying causal graph.
We present a multitude of results pertaining to the identifiability of latent representations, demonstrating these outcomes both empirically and experimentally. Evaluations on synthetic and image datasets corroborate the efficacy of the proposed estimation method, which in practice exhibits superior identifiability. 
Possible future works would be to investigate sample efficiency and robustness of the models trained with the proposed estimation method. Additionally, extending the proposed approach from ANM to post-ANM  and simplifying modelling assumptions  would be of particular interest.
Although modelling assumptions are standard and widely used in practice, formulating a model and estimation methods without these assumptions would be ideal.

\newpage
\bibliographystyle{plain}
\bibliography{neurips_2023}


\newpage
\appendix
\section{Proofs}

\begin{theorem}(Identifiability of $\rvz$ under $\gG$)
Let $\rvf_o, \tilde{\rvf}_o$, satisfying injectivity assumption with $y \sim \mathbb{P}(\rvz), y' \sim  \tilde{\mathbb{P}}(\rvz)$, where $\mathbb{P}, \tilde{\mathbb{P}}$ follow the same causal graph $\gG$. Suppose $\rvf_o(y)$ and $\tilde{\rvf}_o(y')$ are equally distributed, then, $\mathbb{P}(\rvz) \sim \tilde{\mathbb{P}}(\rvz)$.
\end{theorem}

\begin{remark}
This theorem is similar to, but goes beyond, Theorem E.1 in \cite{kivva2022identifiability}. We show equivalence up to $\sim$ rather than $\sim_P$, given that the latent variables are constrained with respect to some causal graph (with all conditional independencies).
\end{remark}

\begin{proof}
The proof is detailed in the Appendix. The main outline of this proof includes showing that, under the constrain that the latent distribution respects the same causal graph $\gG$, the block permutation matrix (in Theorem E.1 of \cite{kivva2022identifiability}) can be reduced to a diagonal matrix. Similar to \cite{kivva2022identifiability}, we approximate the posterior distribution using GMMs.

Based on our formulation, we consider the following:
$$y \sim \mathbb{P}(\rvz) = \prod_i \mathbb{P}_{\mathcal{G}}(\ervz_i \mid \mathbf{pa}(\ervz_i)) =  \sum_{j=0}^{J} \pi(j) \mathcal{N}(\mu_j, \Sigma_j)$$
$$y' \sim \tilde{\mathbb{P}}(\rvz) = \prod_i \tilde{\mathbb{P}}_{\mathcal{G}}(\ervz_i \mid \mathbf{pa}(\ervz_i)) = \sum_{j=0}^{J} \tilde{\pi}(j) \mathcal{N}(\tilde{\mu}_j, \tilde{\Sigma}_j)$$

Now, we consider $y' = Ay + b$; we show that when the latent causal distribution is known, $A$ is the identity matrix.

Without loss of generality, we consider $y'$ to belong to component $k\in \{0, \dots, J\}$. Given that $y' = Ay + b$, a linear transformation of a Gaussian random variable results in:

$$ \tilde{\Sigma}_k = A \Sigma_k A^T$$

As both matrices are diagonal and positive semi-definite (PSD), spectral decomposition using singular value decomposition (SVD) results in $\tilde{\Sigma}_k = V_kV_k^T = {V'}_k{V'}^T_k$, where $V_k, {V'}_k$ are PSD matrices and are unique up to orthogonal transformation $\Rightarrow V_k = R_kV'_k$ for some unitary matrix $R_k$ for each and every $k \in \{0, \dots, J\}$, resulting in:

$$\tilde{\Sigma}_k^{1/2} = V_kR_k = A\Sigma_k^{1/2}$$

Without loss of generality, let's consider two components $k =1$ and $k=2$,
$$\tilde{\Sigma}_1^{-1/2}\Sigma_1^{-1/2} = \tilde{\Sigma}_2^{-1/2}\Sigma_2^{-1/2} \Rightarrow V_1R_1\Sigma_1^{-1/2} = V_2R_2\Sigma_2^{-1/2}$$

By rearranging terms, we get:

$$R_1(\Sigma_1^{-1/2}\Sigma_2^{1/2})R_2^{-1} = V_1^{-1} V_2$$

As $R_1, R_2$ are unitary, $\Sigma_1, \Sigma_2$ are diagonal and PSD, and $y$, $y'$ follow the same causal structure $\mathcal{G}$, the SVD decomposition of $V_1^{-1} V_2$ results in $R'$ such that:

$$V_1R'_1 = A \Sigma_1^{1/2}, \, \mathrm{for}\, A' \coloneqq V_1R_1, \quad \text{we have~} (A')^{-1} A = \Sigma_1^{-1/2} \Rightarrow A=I$$

This concludes the proof.
\end{proof}

\paragraph{Description:} The theorem mainly focuses on showing identifiability results when latent distribution follows the same factorization with respect to a known causal graph and has an injective and a perfect mixing function. Here, we show that based on the fact that GMMs are universal approximators of any arbitrary latent density.

\begin{lemma}(Identifiability of $\rvz$ under causal ordering) In the case when only causal ordering is known, the strong identifiability in theorem \ref{thm:latentidentifiability} reduces to block diagonal identifiability ($\sim_D$).
\end{lemma}

\begin{proof}
Similar to the previous theorem, we consider here the case of
latent variables that can be factorized w.r.t.\ two different causal graphs $\mathcal{G}, \mathcal{G}'$ with the same causal order $\tau$.

$$y \sim \mathbb{P}(\rvz) = \prod_i \mathbb{P}_{\mathcal{G}}(\ervz_i \mid \mathbf{pa}(\ervz_i)) =  \sum_{j=0}^{J} \pi(j) \mathcal{N}(\mu_j, \Sigma_j)$$
$$y' \sim \tilde{\mathbb{P}}(\rvz) = \prod_i \tilde{\mathbb{P}}_{\mathcal{G}'}(\ervz_i \mid \mathbf{pa}(\ervz_i)) = \sum_{j=0}^{J} \tilde{\pi}(j) \mathcal{N}(\tilde{\mu}_j, \tilde{\Sigma}_j)$$

Now, we consider $y' = Ay + b$; we show that when the causal ordering is known, $A$ is a block diagonal matrix.

Without loss of generality, we consider $y'$ to belong to component $k\in \{0, \dots, J\}$. Given that $y' = Ay + b$, a linear transformation of a Gaussian random variable results in:

$$ \tilde{\Sigma}_k = A \Sigma_k A^T$$

As both matrices are diagonal and positive semi-definite (PSD), spectral decomposition using singular value decomposition (SVD) results in $\tilde{\Sigma}_k = V_kV_k^T = {V'}_k{V'}^T_k$, where $V_k, {V'}_k$ are PSD matrices and are unique up to orthogonal transformation $\Rightarrow V_k = R_kV'_k$ for some unitary matrix $R_k$ for each and every $k \in \{0, \dots, J\}$, resulting in:

$$\tilde{\Sigma}_k^{1/2} = V_kR_k = A\Sigma_k^{1/2}$$

Without loss of generality, let's consider two components $k =1$ and $k=2$,
$$\tilde{\Sigma}_1^{-1/2}\Sigma_1^{-1/2} = \tilde{\Sigma}_2^{-1/2}\Sigma_2^{-1/2} \Rightarrow V_1R_1\Sigma_1^{-1/2} = V_2R_2\Sigma_2^{-1/2}$$

By rearranging terms, we get:

$$R_1(\tilde{\Sigma}_2^{1/2}\tilde{\Sigma}_1^{1/2})R_2^{-1} = R_1(\Sigma_1^{-1/2}\Sigma_2^{1/2})R_2^{-1} = V_1^{-1} V_2$$

As $R_1, R_2$ are unitary and $\Sigma_1, \Sigma_2$ are diagonal and PSD with all distinct entries, and $y$ and $y'$ follow same causal order $\tau$, the SVD decomposition of $V_1^{-1} V_2$ results in $R'$, with transformation matrix $\mathbf{D}$ such that:

$$V_1R'_1\mathbf{D} = A \Sigma_1^{1/2}, \, \mathrm{for}\, A' \coloneqq V_1R_1, \quad \text{we have~} (A')^{-1} A = \mathbf{D}\Sigma_1^{-1/2} \Rightarrow A=\mathbf{D}$$

As the entries of $\Sigma_1, \Sigma_2$ are distinct and $y, y'$ follow the same causal order $\tau$, the resulting transformation matrix is either diagonal or block diagonal in nature.

This concludes the proof.

\end{proof}

\paragraph{Description:} Given the fact that constraining latent variables based on the complete causal graph may not be feasible, the lemma relaxes this constraint to enforce causal ordering, which guarantees $\sim_D$ identifiability. Intuitively, the transformation is diagonal in the case when the graph follows a Markov chain structure, and block diagonal when the graph consists of multiple sister nodes at each level.

\begin{theorem}(Model Identifiability)
Let $\rvf_o, \tilde{\rvf}_o$, satisfy the injectivity assumption with $y \sim \mathbb{P}(\rvz), y' \sim  \tilde{\mathbb{P}}(\rvz)$, where $\mathbb{P}, \tilde{\mathbb{P}}$ follow the same causal graph $\gG$ and let $\mathcal{D} \subseteq \mathbb{R}^o$, where $o = |\mathcal{O}|$ such that $\rvf_o, \tilde{\rvf}_o$ are injective on to $\mathcal{D}$. 
Suppose $\rvf_o(y)$ and $\tilde{\rvf}_o(y')$ are equally distributed, then, $\rvf_o(\rvz) = \tilde{\rvf}_o(\rvz)$.

\end{theorem}

\begin{remark}
This theorem is similar to, but goes beyond Theorem D.4 in \cite{kivva2022identifiability}. We show equivalence up to $\sim$ rather than $\sim_A$, given that the latent variables are constrained with respect to some causal graph (with all conditional independencies).
\end{remark}

\begin{proof}
    
We detail the proof in the Appendix. Similar to the proof of Theorem \ref{thm:latentidentifiability}, we use GMMs to model our posterior distribution. 


Similar to the proof of Theorem C.7 in \cite{kivva2022identifiability}, we assume both $\rvf_o, \tilde{\rvf}_o$ are piece-wise affine functions and are invertible on $B(x_0, 2\delta) \cap \rvf_0(\mathbb{R}^d)$, where $B$ is a ball with radius $\delta$. Given both $y, y'$ are sampled from the same causal latent distribution, $\rvf_o \sim \tilde{\rvf}_o$. Since, $\tilde{\rvf}_o$ is invertible on $\mathcal{D}$, $y = (\tilde{\rvf}_o^{-1} \circ \rvf_o)(y)$ on $\rvf_o^{-1}(D)$. This results in $\rvf_o(y) = \tilde{\rvf}_o(y')$ for every $y' \in \tilde{\rvf}_o^{-1}(\mathcal{D})$.

Given that the latent distribution follows the same causal distribution, the injectivity assumption holds, and the mixing functions share the same pre-image, both mixing functions are identifiable.

\end{proof}

\begin{lemma}(Model identifiability under causal ordering) In the case when latent variables follow a particular causal ordering $\tau$ rather than the entire causal graph $\gG$, there exists a block diagonal transformation $\mathbf{D}$ such that $\rvf_o(\rvz) = (\tilde{\rvf}_o \circ \mathbf{D})(\rvz)$.

\end{lemma}

\begin{proof}
Similar to Lemma \ref{lemma:orderedlatentidentifiability}, we consider the latent space to be factorized with respect $\gG, \gG'$ with the topological order $\tau$, resulting in block diagonal transformation function $\mathbf{D}: \mathbb{R}^d \rightarrow \mathbb{R}^d$ such that $x'=\mathbf{D}x$ for $x\sim \gG, x' \sim \gG'$.

As described in the previous theorem, we consider both $\rvf_o, \tilde{\rvf}_o$ are piece-wise affine functions and are invertible on $B(x_0, 2\delta) \cap \rvf_0(\mathbb{R}^d)$.
Based on the transformation $\mathbf{D}$, $\tilde{\rvf}_o (y) \sim (\rvf_o \circ \mathbf{D})(y) \; \forall\; y \in \tilde{\rvf}_o^{-1}(\mathcal{D})$. Since $\rvf_o, \tilde{\rvf}_o$ are invertible, above expression can be rewritten as $y \sim (\tilde{\rvf}_o^{-1} \circ \rvf_o \circ \mathbf{D})(y)$ on $\tilde{\rvf}_o^{-1}(\mathcal{D})$.
In some special cases, where $\mathcal{D}$ is invertible (Markov chain structure in latent space), where $\mathcal{D}$ is mostly diagonal matrix $y \sim (\mathbf{D}^{-1} \circ \rvf_o \circ \rvf_o^{-1} )(y)$ on $\tilde{\rvf}_o^{-1}(\mathcal{D})$, we can infer that $\mathbf{D}$ is a diagonal transformation on $\tilde{\rvf}_o{-1}(\mathcal{D})$.
In general cases, there exists an invertible transformation $\mathbf{D}_1, \mathbf{D}_2$ such that:
$$(\tilde{\rvf}_o^{-1} \circ \rvf_o \circ \mathbf{D}) \sim (\mathbf{D}_1^{-1} \circ \rvf_o \circ \rvf_o^{-1}) + (\mathbf{D}_2^{-1} \circ \rvf_o \circ \rvf_o^{-1})$$

From which we can conclude that $\mathbf{D}_1, \mathbf{D}_2$ are diagonal transformations on $\tilde{\rvf}_o{-1}(\mathcal{D})$, for which we have $\rvf_o(y) = (\tilde{\rvf}_o \circ \mathbf{D})(y') \; \forall y \in \tilde{\rvf}_o^{-1}(\mathcal{D})$.

\end{proof}

\begin{lemma}(Training Objective)
Based on the proposition 1 and lemmas \ref{thm:latentidentifiability}, \ref{thm:modelidentifiability}, and \label{lemma:ELBO_derivation} models trained with the following objective: $\mathcal{L}_{total} = \mathcal{L}_{ELBO} + \alpha \mathcal{L}_{order}$, where 
 will converge at true latents with $\sim_D$ equivalence.   
\end{lemma}

\begin{proof}
ELBO Derivation:

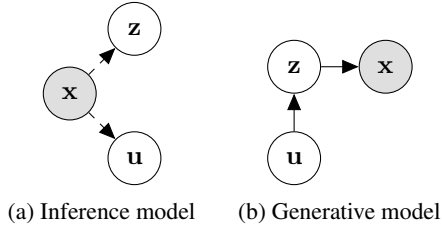
\begin{figure}
    \centering
    \begin{subfigure}{.22\textwidth}
        \centering
        \begin{tikzpicture}
            \node[obs] (x) {$\rvx$};
            \node[latent, above right=0.5cm of x, minimum size=20pt] (w) {$\rvz$};
            \node[latent, below right=0.5cm of x, minimum size=20pt] (u) {$\rvu$};
        
            \path (x) edge [dashed, ->] (w)
            (x) edge [dashed, ->] (u);
        \end{tikzpicture}
    \caption{Inference model}
    \end{subfigure}
    \begin{subfigure}{.22\textwidth}
        \centering
        \begin{tikzpicture}
            \node[obs] (x) {$\rvx$};
            \node[latent, left=0.5cm of x, minimum size=20pt] (w) {$\rvz$};
            \node[latent, below=0.5cm of w, minimum size=20pt] (u) {$\rvu$};
            
            \path (w) edge [->] (x)
            (u) edge [->] (w);
     
        \end{tikzpicture}
        \caption{Generative model}
    \end{subfigure}
    \caption{Variational posterior $\mathbb{Q}(\rvu,\rvz\mid\rvx)$ used during inference on the left and generative model on the right. We do not give a causal interpretation for $\rvc$ in this case. }
    \label{fig:modelling}
\end{figure}

For this, we start with the data distribution as $\mathbb{P}(\rvx)$, and the aim is to maximize the log-likelihood of this distribution:

$$\log \mathbb{P}(\rvx)$$
$$ = \log \int_{\rvu} \int_{\rvz} \mathbb{P}(x, \rvu, \rvz) d\rvz d\rvu$$

Let's consider variational distributions $\mathbb{Q}(\rvu, \rvz \mid \rvx)$.
$$ = \log \int_{\rvu} \int_{\rvz} \mathbb{P}(\rvx, \rvu, \rvz)  \ffrac{\mathbb{Q}(\rvu, \rvz \mid \rvx)}{\mathbb{Q}(\rvu, \rvz \mid \rvx)} d\rvz d\rvu$$

$$ \geq \E_{\mathbb{Q}(\rvu, \rvz \mid \rvx)} \log  \ffrac{\mathbb{P}(\rvx, \rvu, \rvz)}{\mathbb{Q}(\rvu, \rvz \mid \rvx)} $$

Based on modelling assumption described in figure \ref{fig:modelling}, $\mathbb{Q}(\rvu, \rvz \mid \rvx)$ decomposes as $\mathbb{Q}(\rvu \mid \rvx)\mathbb{Q}(\rvz \mid \rvx)$

$$ = \E_{\mathbb{Q}(\rvu, \rvz \mid \rvx)} \Big[ \log \mathbb{P}(\rvx \mid\rvz) + 
\log \ffrac{\mathbb{P}(\rvu \mid\rvz)}{\mathbb{Q}(\rvu \mid \rvx)} + 
\log \ffrac{\mathbb{P}(\rvz)}{\mathbb{Q}(\rvz \mid \rvx)} \Big]$$

$$ = \E_{\mathbb{Q}(\rvz \mid \rvx)} \log \mathbb{P}(\rvx \mid\rvz) + 
\E_{\mathbb{Q}(\rvz \mid \rvx)} \E_{\mathbb{Q}(\rvu\mid \rvx)}\log \ffrac{\mathbb{P}(\rvu \mid\rvz)}{\mathbb{Q}(\rvu \mid \rvx)} + 
\E_{\mathbb{Q}(\rvz \mid \rvx)} \log \ffrac{\mathbb{P}(\rvz)}{\mathbb{Q}(\rvz \mid \rvx)} $$

$$ = \E_{\mathbb{Q}(\rvz \mid \rvx)} \log \mathbb{P}(\rvx \mid\rvz) - 
\E_{\mathbb{Q}(\rvz \mid \rvx)} \mathrm{KL}\Big(\mathbb{P}(\rvu \mid\rvz) \|\mathbb{Q}(\rvu \mid \rvx)\Big) - 
\mathrm{KL}\Big(\mathbb{P}(\rvz)\| \mathbb{Q}(\rvz \mid \rvx)\Big)$$

$$\Rightarrow \mathcal{L}_{ELBO} = - \E_{\mathbb{Q}(\rvz \mid \rvx)} \log \mathbb{P}(\rvx \mid\rvz) +
\E_{\mathbb{Q}(\rvz \mid \rvx)} \mathrm{KL}\Big(\mathbb{P}(\rvu \mid\rvz) \|\mathbb{Q}(\rvu \mid \rvx)\Big) +
\mathrm{KL}\Big(\mathbb{P}(\rvz)\| \mathbb{Q}(\rvz \mid \rvx)\Big)$$

Based on proposition 1, we can infer that as $\mathcal{L}_{order} \rightarrow 0$ the causal order is enforced in latent space:
$$\mathbb{P}(\rvz) = \prod_i \mathbb{P}(\ervz_i\mid \{\ervz_{i+1}, \dots \ervz_d\})$$

Based on our assumptions, the considered model is injective, and based on lemma \ref{lemma:orderedlatentidentifiability} and \ref{lemma:orderedmodelidentifiability} we know that the latent distribution and model converges to a unique solution with $\sim_D$ equivalence given the causal ordering of latent space.

Given infinite training data and compute, as $\mathcal{L}_{total} \rightarrow 0$, $\mathcal{L}_{ELBO} \rightarrow 0$ converging the obtained unique distribution to true prior distribution.
\end{proof}

\section{Data Generating Process}


\subsection{\textsc{MorphoMNIST} dataset}

Here, we synthetic data based on MNIST digits \cite{castro2019morpho}. 
We define multiple data-generating process with four different variables thickness, width, slant, and intensity, and evaluate our proposed method in terms of MCC's and COD.
Here, thickness corresponds to the stroke thickness of a digit, width corresponds to the total width of a written digit, slant corresponds to the shear factor along a horizontal direction, and intensity corresponds to the average intensity of pixels in a digit.
Functions $SetIntensity( x; i)$, $SetSlant( x; s)$, $SetWidth( x; w)$, and $SetThickness( x; t)$ refer to the operations applied to the original MNIST digit to generate new image $x$ with desired properties by controlling image morphology.
We use the data-generating process similar to the ones described in \cite{glance}, we formally describe them below.

\textbf{Morpho-MNIST-TI}: In this setting we consider two causal variables thickness and intensity, where thickness causes intensity. 
Mathematically the functional relationship between variables are defined as described in equation \ref{eqn:morphoMNISTTIEqns}.

\begin{equation}
    \begin{aligned}
        &t := f_t \triangleq 0.5 + \epsilon_t \quad \epsilon_t \sim \Gamma(10, 5) \\
        &i := f_i \triangleq 64 + 191*\sigma(2*w + 5) + \epsilon_i \quad \epsilon_i \sim \mathcal{N}(0, 1)\\
        &x := f_x = SetIntensity(SetThickness(X; t); i)
    \end{aligned}
    \label{eqn:morphoMNISTTIEqns}
    \centering
\end{equation}

\textbf{Morpho-MNIST-IT}: In this experiment we inverted a directionality from previous setting resulting in intensity to cause thickness, which is mathematically described in equation \ref{eqn:morphoMNISTITEqns}

\begin{equation}
    \begin{aligned}
        &i := f_i \triangleq \epsilon_i \quad \epsilon_i \sim \mathbb{U}(60, 255)\\
        &t := f_t \triangleq 3 + \sigma(i/255) + \epsilon_s \quad \epsilon_s \sim \mathcal{N}(0, 0.5)\\
        &x := f_x = SetThickness(SetIntensity(X; i); t)
    \end{aligned}
    \label{eqn:morphoMNISTITEqns}
    \centering
\end{equation}

\textbf{Morpho-MNIST-TS}: In this setup we use thickness and slant as causal attributes, where thickness causes digit slantness, which is formally described in equation \ref{eqn:morphoMNISTTSEqns}

\begin{equation}
    \begin{aligned}
        &t := f_t \triangleq \epsilon_t \quad \epsilon_t \sim \Gamma(0, 5) \\
        &s := f_s \triangleq 10 + 5*\sigma(2*t - 5) + \epsilon_s \quad \epsilon_s \sim \mathcal{N}(0, 0.5)\\
        &x := f_x = SetSlant(SetThickness(X; t); s)
    \end{aligned}
    \label{eqn:morphoMNISTTSEqns}
    \centering
\end{equation}

\textbf{Morpho-MNIST-TSWI}: In this setup we increased a complexity by using intensity, thickness, slant, and digit width as a causal attributes, where thickness causes slant, thickness and slant causes width, and width causes intensity.
This data-generating process is formally described in equation \ref{eqn:morphoMNISTTSWIEqns}

\begin{equation}
    \begin{aligned}
        &t := f_t \triangleq \epsilon_t \quad \epsilon_t \sim \Gamma(0, 5) \\
        &s := f_s \triangleq 10 + 20*t + \epsilon_s \quad \epsilon_s \sim \mathcal{N}(0, 5)\\
        &w := f_w \triangleq  10 + 15*\sigma(0.5*t) - 0.25*s  + \epsilon_w
     \quad \epsilon_w \sim \mathcal{N}(0, 1) \\
        &i := f_i \triangleq 64 + 191*\sigma(w/25) + \epsilon_i \quad \epsilon_i \sim \mathbb{N}(0, 1)\\
        &x := f_x = SetIntensity(SetWidth(SetSlant(SetThickness(X; t); s); w); i)
    \end{aligned}
    \label{eqn:morphoMNISTTSWIEqns}
    \centering
\end{equation}

\section{Experimental Setup}

\subsection{Code and Implementation}

We use the latent GMM loss from MFC-VAE \cite{falck2021multi} inspired in the implementation from \url{https://github.com/FabianFalck/mfcvae}. We also append the code for the model and loss functions used in the paper to the supplemental material.

\subsection{Hyperparameters}
In Table \ref{tab:hparams} we detail all the hyper-parameters used in our experiments. We use a fixed decoder standard deviation in the case of \textsc{Causal3DIdent} and \textsc{MorphoMNIST}, while in the case of \textsc{Syn-k} dataset it remains learnable (described as $\sigma$ in the table). It is also worth mentioning that for the VAE method on \textsc{Causal3DIdent}, we trained a deeper model and also set the KL weight term $\beta$ equal to 0 to ensure fair comparison with the other two methods and avoid posterior collapse, respectively.

\begin{table}
    \caption{Experimental details w.r.t models and datasets}
    \centering
        \begin{tabular}{@{}llccc@{}}
            \toprule
            \multirow{2}{*}{\begin{tabular}[c]{@{}l@{}} \textsc{Datasets}($\downarrow$),\\ \textsc{Methods}($\rightarrow$)\end{tabular}} \\
                       &  & VAE  & MFC-VAE   & \textsc{coVAE} \\ \cmidrule(r){1-5}
            \multirow{9}{*}{\textsc{Syn-k}}
             & No. Layers & \multicolumn{3}{c}{3 if k < 3 else 6}\\
             & Training Steps  & \multicolumn{3}{c}{15600}\\
             & No. Samples  & \multicolumn{3}{c}{2000}\\
             & Batch Size & \multicolumn{3}{c}{256}\\
             & Optimizer & \multicolumn{3}{c}{Adam}\\
             & Learning Rate & \multicolumn{3}{c}{5e-4}\\
             & $\alpha$ & - & 0.0 & 1.0\\
             & $\beta$ & 1.0 & 1.0 & 1.0\\
             & Decoder $\sigma$ & \multicolumn{3}{c}{$\sigma$}\\ \cmidrule(r){1-5}
             
             \multirow{9}{*}{\textsc{MorphoMNIST}}
             & No. Layers & \multicolumn{3}{c}{6}\\
             & Training Steps  & \multicolumn{3}{c}{~6000}\\
             & No. Samples  & \multicolumn{3}{c}{60000}\\
             & Batch Size & \multicolumn{3}{c}{256}\\
             & Optimizer & \multicolumn{3}{c}{Adam}\\
             & Learning Rate & \multicolumn{3}{c}{1e-4}\\
             & $\alpha$ & - & 0.0 & 1.0\\
             & $\beta$ & 1.0 & 1.0 & 1.0\\
             & Decoder $\sigma$ & 0.5 & 0.5 & 0.5 \\ \cmidrule(r){1-5}
             
             \multirow{9}{*}{\textsc{Causal3DIdent}}
             & Input resolution & \multicolumn{3}{c}{$64\times64$} \\
             & No. Layers & 4 & 3 & 3 \\
             & Training Steps  & \multicolumn{3}{c}{19687} \\
             & No. Samples  & \multicolumn{3}{c}{252000}\\
             & Batch Size & \multicolumn{3}{c}{128} \\
             & Optimizer & \multicolumn{3}{c}{Adam}\\
             & Learning Rate & \multicolumn{3}{c}{5e-4} \\
             & Hidden dim & \multicolumn{3}{c}{256} \\
             & Latent dim & 256 & 16 & 16 \\
             & $\alpha$ & - & 1.0 & 1.0 \\
             & $\beta$  & 0.0 & 0.01 & 0.01 \\
             & Decoder $\sigma$ & 0.1 & 0.1 & 0.1 \\
             \bottomrule
        \end{tabular}
    \label{tab:hparams}
\end{table}

\section{Results}
Table \ref{tab:results} depicts final results on \textsc{MorphoMNIST-TI}, \textsc{MorphoMNIST-TS}, and \textsc{Causal3DIdent} dataset, respectively. For each method, we re-run all experiments and collect metrics across 5 different random seeds for \textsc{MorphoMNIST-TI} and \textsc{MorphoMNIST-TS}, and 3 random seeds for \textsc{Causal3DIdent}. For the latter dataset, we observed that all three metrics exhibit high variance across runs; however, it is clear that both MFC-VAE and \textsc{coVAE} are comparable methods. 

\begin{table}
    \caption{MCC and COD results on MorphoMNIST and Causal3DIdent datasets}
    \centering
        \begin{tabular}{@{}cccc@{}}
            \toprule
            \multirow{2}{*}{\begin{tabular}[c]{@{}l@{}} \textsc{Methods}($\downarrow$),\\ \textsc{Metrics}($\rightarrow$)\end{tabular}} & \multicolumn{3}{c}
             {\textsc{MorphoMNIST-TI}} \\\cmidrule(l){2-4}
             & COD ($\downarrow$) & MCC-R($\uparrow$)   & MCC-SG($\uparrow$) \\ \cmidrule(r){1-4}
             VAE &  1.31 $\pm$ 0.28 & 0.31 &  0.24 $\pm$ 0.06\\
             MFC-VAE & 1.33 $\pm$ 0.38  & 0.38 & \textbf{0.39} $\pm$ 0.07\\
             \textsc{coVAE} & \textbf{0.0}   & \textbf{0.58} &  0.38 $\pm$ 0.06\\ \cmidrule(r){1-4}
             
             & \multicolumn{3}{c}{\textsc{MorphoMNIST-TS}} \\ \cmidrule(r){1-4}
             VAE  & 1.47 $\pm$ 0.65    & 0.48 & 0.38 $\pm$ 0.05\\
             MFC-VAE & 1.75 $\pm$ 0.60 & 0.51 & 0.36 $\pm$ 0.06\\
             \textsc{coVAE} & \textbf{0.0}      & \textbf{0.56} & \textbf{0.41} $\pm$ 0.05 \\ \cmidrule(r){1-4}
             & \multicolumn{3}{c}{\textsc{Causal3DIdent}} \\ \cmidrule(r){1-4}
             VAE & 22.39 $\pm$ 1.49 & 0.15 & 0.15 $\pm$ 0.0 \\
             MFC-VAE & \textbf{3.56} $\pm$ 0.87 & \textbf{0.28} & \textbf{0.27} $\pm$ 0.01 \\
             \textsc{coVAE} & 3.94 $\pm$ 0.86 & 0.26 & 0.25 $\pm$ 0.02 \\
             \bottomrule
        \end{tabular}
    \label{tab:results_additional}
\end{table}

\end{document}